\documentclass[runningheads]{llncs}
\pdfoutput=1

\usepackage{amsmath,amssymb}
\usepackage{times,helvet,courier}
\usepackage{url}
\usepackage{multirow}

\usepackage[listofformat=parens,subrefformat=subparens]{subfig}

\usepackage{listings}
\newcommand{\quontroller}{\textit{quontroller}}

\title{A System for Interactive Query Answering\\ with Answer Set Programming}

\author{Martin Gebser \and Philipp Obermeier \and Torsten Schaub\thanks{Affiliated with the
                              School of Computing Science at
                              Simon Fraser University,
                              Burnaby, Canada,
                              and the
                              Institute for Integrated and Intelligent Systems
                              at
                              Griffith University,
                              Brisbane, Australia.}}

\institute{Universit\"at Potsdam, Institut f\"ur Informatik}

\titlerunning{A System for Interactive Query Answering with Answer Set Programming}
\authorrunning{M.~Gebser \emph{et al}\/.} 
\begin{document}

\maketitle

\begin{abstract}
Reactive answer set programming has paved the way for incorporating online information into
operative solving processes.
Although this technology was originally devised for dealing with data streams in dynamic environments,
like assisted living and cognitive robotics,
it can likewise be used to incorporate facts, rules, or queries provided by a user.
As a result,
we present the design and implementation of 
a system for interactive query answering with reactive answer set programming.
Our system \textit{quontroller} is based on the reactive solver \textit{oclingo}
and implemented as a dedicated front-end.
We describe its functionality and implementation,
and we illustrate its features by some selected use cases.
\end{abstract}


\section{Introduction}\label{sec:introduction}

Traditional logic programming \cite{clomel81,lloyd87} is based upon query answering.
Unlike this,
logic programs under the stable model semantics \cite{gellif88b} are implemented by model generation based systems,
viz.\ answer set solvers~\cite{gekakasc12a}.
Although the latter also allows for checking whether a query is entailed by some stable model,
there is so far no way to explore a domain at hand by posing consecutive queries without relaunching the solver.
The same applies to the interactive addition and/or deletion of temporary program parts that come in handy
during theory exploration, for instance, when dealing with hypotheses.

An exemplary area where such exploration capacities would be of great benefit is bio-informatics
(cf.\ \cite{bachtrtrjobe04a,erdtur08a,gescthve10a,geguivscsithve10a,rawhki10a,viguedthgrsasi12a}).
Here, we usually encounter problems with large amounts of data, 
resulting in runs having substantial grounding and solving times.
Furthermore, 
problems are often under-constrained, thus yielding numerous alternative
solutions.
In such a setting, it would be highly beneficial to explore a domain via successive queries and/or
under certain hypotheses.
For instance,
for determining nutritional requirements for sustaining maintenance or growth of an organism,
it is important to indicate seed compounds needed for the synthesis of 
other compounds.
Now, rather than continuously analyzing several thousand stable models (or their intersection or
union), a biologist may rather perform interactive ``in-silico'' experiments by temporarily adding 
compounds and subsequently exploring the resulting models by posing successive queries.

We address this shortcoming and show how recently developed systems for 
\emph{reactive} answer set programming (ASP) \cite{gegrkasc11a,gegrkaobsasc12b} can be harnessed to
provide query answering and theory exploration capacities.
In fact, reactive ASP was conceived for incorporating online information into operative ASP solving processes.
Although this technology was originally devised for dealing with data streams in dynamic environments,
like assisted living and cognitive robotics,
it can likewise be used to incorporate facts, rules, or queries provided by a user.
As a result,
we present the design and implementation of a system for interactive query answering and theory
exploration with ASP.
Our system \textit{quontroller}\footnote{%
To be pronounced `cointreau'-ler; 
URL: \url{potassco.sourceforge.net/labs.html}.}
is based on the reactive answer set solver \textit{oclingo}
and implemented as a dedicated front-end.
We describe its functionality and implementation,
and we illustrate its features 
on a running example.
%


\section{Approach}\label{sec:approach}

In order to provide dedicated support for query answering,
the \textit{quontroller}
encapsulates 
\textit{oclingo} along with its
basic front-end
for entering online progressions.
The basic idea is to condition the stable models of an underlying logic program
via \emph{query programs},
temporarily asserting atoms to be contained in stable models of interest.
For circumventing restrictions due to the modularity requirement
of reactive ASP (cf.\ \cite{gegrkasc11a,gegrkaobsasc12b})
and enabling repeated assertions of an atom in a series of queries,
the \textit{quontroller} associates query programs with sequence numbers and
exploits \textit{oclingo}'s step counter to automatically map their contents.
In the following, we detail this idea on the well-known example of $n$-coloring.

To begin with,
Listing~\ref{lst:encoding} provides an ASP encoding of $n$-coloring.
The encoding applies to graphs represented by facts over predicate \lstinline{edge}/2.
Given such facts,
the nodes of the graph are extracted in Line~6--7, each node is
marked with exactly one    of~\lstinline{n} colors in Line~10,
and   Line~11 forbids  that connected nodes are marked with the same color.
In Line~14--15, stable models are projected onto atoms over predicate \lstinline{mark}/2.
%
\lstinputlisting[%
float=t,frame=single,basicstyle=\ttfamily\footnotesize,
caption={ASP encoding of $n$-coloring (\lstinline{encoding.lp})},%
label=lst:encoding]%
{encoding.lp}
\lstinputlisting[%
float=t,frame=single,basicstyle=\ttfamily\footnotesize,
caption={\textit{quontroller} setup instructions (\lstinline{setup.ini})},%
label=lst:setup]%
{setup.ini}
\lstinputlisting[%
float=t,frame=single,basicstyle=\ttfamily\scriptsize,
caption={Mapping of \textit{quontroller} setup instructions in 
\lstinline{setup.ini} to reactive ASP rules},%
label=lst:reactive]%
{setup.lp}

Unlike with    one-shot ASP solving,
we do not combine the encoding in Listing~\ref{lst:encoding} with fixed facts,
but rather aim at a selective addition as well as withdrawal of atoms over \lstinline{edge}/2.
In order to prepare reactive ASP rules for this purpose,
the instructions in Listing~\ref{lst:setup} can be fed into the \textit{quontroller}.
They are then mapped to the language of \textit{oclingo} as shown
in Listing~\ref{lst:reactive}.
The resulting reactive ASP rules are divided into
a static \lstinline{#base} part (Line~1--13),
a stepwise \lstinline{#cumulative} part (Line~15--25), and
a stepwise \lstinline{#volatile} part (Line~27--34).
The stepwise parts are instantiated for successive step numbers
replacing the constant~\lstinline{t},
where instances of \lstinline{#cumulative} rules are gathered over steps and
\lstinline{#volatile} ones are discarded when progressing to the next step.

In more detail,
the \lstinline{#domain} instructions in Line~2 and~3 of Listing~\ref{lst:setup}
are mapped to the static rules in Line~4 and~7 of Listing~\ref{lst:reactive}.
They define the (\textit{quontroller}-internal) predicates
\lstinline{_domain_edge}/2 and \lstinline{_domain_mark}/2,
which provide the domains of instances of \lstinline{edge}/2 and
\lstinline{mark}/2 that can be asserted by query programs.
In particular, the rule in Line~4 expresses that edges may connect
distinct nodes with labels running from~\lstinline{1} to~\lstinline{4},
and the rule in Line~7 declares that each node can be marked
with the colors provided by \lstinline{color}/1.
Furthermore, the instruction in Line~4 of Listing~\ref{lst:setup} is mapped to
the static choice rule in Line~10 of Listing~\ref{lst:reactive}.
This rule  compensates for the lack of facts over \lstinline{edge}/2 and allows
for instantiating the original encoding in Listing~\ref{lst:encoding}
relative to the domain given by \lstinline{_domain_edge}/2.
Again relying on \lstinline{_domain_edge}/2,
the instruction in Line~7 of Listing~\ref{lst:setup} is mapped to
the \lstinline{#show} statement in Line~13 of Listing~\ref{lst:reactive} for
including atoms over \lstinline{edge}/2
(in addition to those over \lstinline{mark}/2) in output projections.

The \lstinline{#cumulative} and \lstinline{#volatile} parts
in Listing~\ref{lst:reactive} deal with the instructions in Line~5 and~6
of Listing~\ref{lst:setup}.
In particular, the \lstinline{#external} statements in Line~18 and~23 of
Listing~\ref{lst:reactive} declare (\textit{quontroller}-internal) instances of 
\lstinline{_assert_mark(X1,X2,t)} and \lstinline{_assert_edge(X1,X2,t)} as
potential online inputs from query programs,
where~\lstinline{X1} and~\lstinline{X2} are instantiated relative to domains
given by static rules and the constant~\lstinline{t} is added as an argument
to distinguish separate queries.
The rules in Line 19--20 and 24--25 of Listing~\ref{lst:reactive} further
define the (\textit{quontroller}-internal) predicates
\lstinline{_derive_mark}/3 and \lstinline{_derive_edge}/3 for
indicating ``active'' assertions from query programs.
Given that such assertions may remain active over several queries,
the rules defining \lstinline{_derive_mark}/3 and \lstinline{_derive_edge}/3
include one case for reflecting current assertions (Line~19 and~24)
and another for passing on former assertions (Line~20 and~25).
As a consequence,
instances of \lstinline{_derive_mark(X1,X2,t)} and \lstinline{_derive_edge(X1,X2,t)}
capture active assertions regarding the original predicates
\lstinline{mark}/2 and \lstinline{edge}/2,
and corresponding matches are established via \lstinline{#volatile} integrity constraints.
For one, the \lstinline{#query} instruction in Line~6 of Listing~\ref{lst:setup} is mapped to
the integrity constraint in Line~30 of Listing~\ref{lst:reactive},
thus requiring \lstinline{mark(X1,X2)} to hold at any step where
\lstinline{_derive_mark(X1,X2,t)} indicates an active assertion,
and the analogous integrity constraint in Line~33 is obtained in view of the
\lstinline{#define} instruction in Line~5 of Listing~\ref{lst:setup}.
The latter is complemented by another integrity constraint in Line~34,
denying \lstinline{edge(X1,X2)} to hold when \lstinline{_derive_edge(X1,X2,t)}
does not indicate any active assertion.
That is, a \lstinline{#query} instruction expresses that assertions may
require atoms to belong to stable models of interest,
and a \lstinline{#define} instruction is stronger by additionally claiming
some active assertion for atoms to hold.

After launching \textit{oclingo} with the encoding in Listing~\ref{lst:encoding} and 
the reactive ASP rules in Listing~\ref{lst:reactive}
(via `\lstinline{quontroller.py -o encoding.lp -c setup.ini}'),
the \textit{quontroller} is ready to process query programs provided by a user.
An exemplary stream of query programs is shown in Figure~\subref{lst:queries},
and Figure~\subref{lst:queried} provides its counterpart
in the syntax of \textit{oclingo}'s basic front-end.
In fact, the \textit{quontroller} maps query programs to available stream constructs
and automatically performs replacements for interacting with reactive ASP rules.
To begin with,
the keywords `\lstinline{#query.}' and `\lstinline{#endquery.}',
which encapsulate individual query programs, are mapped to
`\lstinline{#step q : 0. #forget q-1.}' and `\lstinline{#endstep.}',
where \lstinline{q} is the sequence number of a query program and
the \lstinline{#forget} directive enables simplifications of reactive ASP rules
for yet undefined \lstinline{#external} atoms introduced at step \lstinline{q-1}.
Also note that `\lstinline{: 0}' in \lstinline{#step} directives
tells \textit{oclingo} not to increment the step counter on
unsatisfiability.

\begin{figure}[t]
\subfloat[\textit{quontroller} query stream\label{lst:queries}]{%
\begin{minipage}{0.4\textwidth}
\begin{lstlisting}[frame=single,basicstyle=\ttfamily\footnotesize]
#query.
#assert : e(1).
edge(1,2).
edge(1,3).
edge(2,3).
edge(2,4).
edge(3,4).
#endquery.

#query.
#assert.
mark(1,1).
#endquery.

#query.
#assert : e(2).
edge(1,4).
#endquery.

#query.
#retract : e(2).
#assert.
mark(1,1).
mark(2,2).
#endquery.

#query.
#retract : e(1).
#endquery.

#stop.
\end{lstlisting}\end{minipage}}
\hfill
\subfloat[Mapping of \textit{quontroller} query stream in~\protect\subref*{lst:queries} \label{lst:queried}]{%
\begin{minipage}{0.48\textwidth}
\begin{lstlisting}[frame=single,basicstyle=\ttfamily\footnotesize]
#step 1 : 0.  #forget 0.
#assert : e(1).
_assert_edge(1,2,1).
_assert_edge(1,3,1).
_assert_edge(2,3,1).
_assert_edge(2,4,1).
_assert_edge(3,4,1).
#endstep.

#step 2 : 0.  #forget 1.
#volatile : 1.
_assert_mark(1,1,2).
#endstep.

#step 3 : 0.  #forget 2.
#assert : e(2).
_assert_edge(1,4,3).
#endstep.

#step 4 : 0.  #forget 3.
#retract : e(2).
#volatile : 1.
_assert_mark(1,1,4).
_assert_mark(2,2,4).
#endstep.

#step 5 : 0.  #forget 4.
#retract : e(1).
#endstep.

#stop.
\end{lstlisting}\end{minipage}}
\vspace{-5.15mm}
\caption{\label{fig:stream}A \textit{quontroller} query stream and its mapping to a reactive ASP online progression}
\end{figure}
Each query program may include labeled assertions,
as declared via `\lstinline{#assert : e(1).}' in Line~2 of
Figure~\subref{lst:queries} and~\subref{lst:queried}.
Such a construct expresses that subsequently provided rules remain active until
the labeled assertion is explicitly retracted.
In view of its reactive ASP rules, the \textit{quontroller} however replaces
each head atom \lstinline{p(...)} of a rule (or fact) to assert by its
internal representation
\lstinline{_assert_p(...,q)}, where \lstinline{q} is again the sequence number
of the query program at hand.
For instance, `\lstinline{edge(1,2).}' is mapped to `\lstinline{_assert_edge(1,2,1).}'
in Line~3 of Figure~\subref{lst:queries} and~\subref{lst:queried}.
By means of reactive ASP rules capturing the \lstinline{#define} instruction
in Line~5 of Listing~\ref{lst:setup},
the instances of \lstinline{_assert_edge}/3 provided by facts in Line~3--7
of Figure~\subref{lst:queried} are matched with the original atoms
over~\lstinline{edge}/2.
As a consequence, the first query program yields six stable models,
in which the unconnected nodes~\lstinline{1} and~\lstinline{4} share one
of the colors~\lstinline{1}, \lstinline{2}, or~\lstinline{3} and the
nodes~\lstinline{2} and~\lstinline{3} are marked with distinct remaining colors.

The second query program in Line~10--13 of Figure~\subref{lst:queries}
includes `\lstinline{mark(1,1).}' as an unlabeled assertion,
indicated by the keyword `\lstinline{#assert.}'
The latter is mapped to the stream construct
`\lstinline{#volatile : 1.}' (cf.\ Line~11 of Figure~\subref{lst:queried}),
meaning that the internal representation
`\lstinline{_assert_mark(1,1,2).}' of the assertion
expires ``automatically'' in the next step.
Technically, such expiration is implemented by adding
assumption literals to the bodies of transient rules, i.e.\
`\lstinline{_assert_mark(1,1,2).}' is internally turned into
`\lstinline{_assert_mark(1,1,2) :- _expire(3).}' and
\lstinline{_expire(3)} holds up to step~\lstinline{3} where it is then
permanently falsified.
However, in the second step, the assertion of color~\lstinline{1} for
node~\lstinline{1} leads to two stable models of interest among
the six obtained in the first step.
Also note that the \textit{quontroller} language includes `\lstinline{#assert.}'
in order to indicate the beginning of query parts in which head atoms are
replaced by internal representations, so that any rules to be left untouched
can still be provided beforehand.

Summarizing the remaining query programs,
the labeled assertion \lstinline{e(2)} in the third query program turns the graph
represented by atoms over \lstinline{edge}/2 into a clique of
four nodes, so that no stable model is obtained in the third step.
Hence, \lstinline{e(2)} is retracted in the fourth step
(by discharging an assumption literal associated with \lstinline{e(2)}),
and the additional unlabeled assertion of colors for the
nodes~\lstinline{1} and~\lstinline{2} leads to a single stable model of interest.
Note that `\lstinline{mark(1,1).}' is turned into `\lstinline{_assert_mark(1,1,4).}'
in Line~23 of Figure
~\subref{lst:queried},
while `\lstinline{_assert_mark(1,1,2).}' has been used in Line~12.
The rewriting by the \textit{quontroller} thus avoids a clash with
\textit{oclingo}'s modularity requirement and enables repeated assertions
of the ``same'' atom (in separate query programs).
Finally, the empty (projection of a) stable model is obtained after
retracting all instances of \lstinline{_assert_edge}/3 in the last query program,
and `\lstinline{#stop.}' afterwards signals the end of the query stream to the \textit{quontroller}.

\section{Discussion}\label{sec:discussion}

We presented a simple yet effective extension of reactive ASP that allows for
interactive query answering and theory exploration with ASP.
This was accomplished by means of a mapping scheme between queries and reactive ASP rules
along with
the assumption-based solving capacities of \textit{oclingo}.
With it, programs can be temporarily added to the solving process,
either for an initially limited number of interactions or until they are interactively withdrawn again.
A typical use case of limited program parts are integrity constraints,
representing queries automatically vanishing after having been posed.
Unlike this, an assertion allows, for instance, for exploring the underlying domain under user-defined
hypotheses.
All subsequent solving processes then include the asserted information until it is retracted by the user.
The possibility of reusing ground rules as well as recorded conflict information over a sequence
of queries distinguishes reactive ASP from ordinary one-shot reasoning methods.
As future work, we want to study
the performance of query answering with the \quontroller\ on
challenging benchmark problems.
%


\paragraph{Acknowledgments}

This work was partially funded by DFG grant SCHA 550/9-1.
We are grateful to the anonymous reviewers for their suggestions.




\end{document}